\documentclass{article}                
\usepackage{spconf,amsmath,epsfig}

\usepackage[utf8]{inputenc} 
\usepackage[T1]{fontenc}    
\usepackage[hidelinks]{hyperref}  
\usepackage{url}            
\usepackage{booktabs}       
\usepackage{amsfonts}       
\usepackage{nicefrac}       
\usepackage{microtype}      
 
\usepackage{graphicx}                                
\usepackage{amssymb}                            
\usepackage{booktabs}                                       
\usepackage{algorithm}                              
\usepackage{algorithmic}   
\usepackage{xcolor}

\usepackage{adjustbox}

\title{Learning from Unlabelled Data with Transformers: Domain Adaptation for Semantic Segmentation of High Resolution Aerial Images} 
\name{Nikolaos Dionelis$^1$, Francesco Pro$^2$, Luca Maiano$^2$, Irene Amerini$^2$, Bertrand Le Saux$^1$}   
\address{$^1$ European Space Agency (ESA), ESRIN, $\Phi$-lab, Italy \\ $^2$Sapienza University of Rome, Italy}  
\begin{document}                        
%
\maketitle          
\begin{abstract}       
Data from satellites or aerial vehicles are most of the times unlabelled. Annotating such data accurately is difficult, requires expertise, and is costly in terms of time. Even if Earth Observation (EO) data were correctly labelled, labels might change over time. Learning from unlabelled data within a semi-supervised learning framework for segmentation of aerial images is challenging. In this paper, we develop a new model for semantic segmentation of unlabelled images, the Non-annotated Earth Observation Semantic Segmentation (NEOS) model. NEOS performs domain adaptation as the target domain does not have ground truth masks. The distribution inconsistencies between the target and source domains are due to differences in acquisition scenes, environment conditions, sensors, and times. Our model aligns the learned representations of the different domains to make them coincide. The evaluation results show that it is successful and outperforms other models for semantic segmentation of unlabelled data.
\end{abstract}
\section{Introduction}     
\label{sec:intro}

\textbf{Importance and overview.}         
Remote Sensing (RS) images from satellites or aerial vehicles can be used to map trees and land cover classes \cite{ModelDIAL}. 
While both RS technology and AI for data analysis continue to advance~\cite{tuia2023artificial}, the integration and use of airplanes and drones for localized studies is nowadays also increasing. 
Supervised learning has shown good performance for classification and segmentation. 
However, it requires high-quality handcrafted \textit{large} labelled datasets. 
Learning from unlabelled data is challenging as the performance of models depends highly on the \textit{size} and quality of the data. 
However, for real-world applications \cite{ChallengeIEEE}, labelling large datasets is laborious, expensive, and time-consuming. 
This holds for Earth Observation (EO), where huge amounts of data are produced \textit{daily}. 
Also, data from satellites or aircrafts usually require domain expertise.
Furthermore, labels for specific geographical regions may change over time (task of change detection) due to nature (seasonality), man-induced changes, and natural hazards (volcano eruptions). 
Also, for specific regions, some labels might be incorrect (task of learning from \textit{noisy labels}). 
Because many satellites and aerial images are unlabelled, it is challenging to effectively use these data. 
Developing semi-supervised learning methods is crucial to improve generalization performance. 
Semi-supervised learning, which involves training on both a labelled dataset, where both images and their annotations are provided, and on an \textit{unlabelled} set, with only image data, is a more realistic setting than supervised learning, as in RS, unlabelled data are \textit{plentiful}, while labelled data can be hard to find. 
This holds for semantic segmentation (\textit{pixel-level} labels) \cite{ModelDIAL},
which requires assigning a class label to each pixel \cite{DeepInterpretable, SemiSupervisedMiniFrance} by understanding its semantics. This task is crucial for several applications, including land cover mapping and urban change detection. 
In this work, we propose a method to perform semantic segmentation on unlabelled datasets, and we evaluate it on the unlabelled Cross-View USA (CVUSA) dataset \cite{CVUSAPaper}. 
To the best of the authors' knowledge, \textit{accurate} semantic segmentation on the CVUSA aerial dataset, which is used for cross-view aerial-ground matching \cite{WhereLooking, SamePlace} and has no land cover label annotations, has not yet been performed.

\textbf{Domain adaptation.}                                                  
In this paper, we develop a model for semantic segmentation of aerial images, the Non-annotated Earth Observation Semantic Segmentation (NEOS) model. 
NEOS makes the learned representations of the \textit{different domains} to coincide. 
This is achieved by minimizing the distribution differences of the different domains. 
Hence, we enforce the model to be able to work well with the different datasets that have distribution inconsistencies due to differences in acquisition scenes, environment conditions, sensors, and times. 
Our model performs semantic segmentation on the unlabelled dataset CVUSA. 
During training, a loss function is minimized that makes the network to align the latent features of the \textit{different domains} to minimize distribution differences. 
Our main contribution is the development of a novel model for semantic segmentation of aerial images that do \textit{not} have ground truth segmentation masks, also performing domain adaptation.

\section{Related Work}  
\label{sec:format}


\textbf{Domain adaptation}               
methods in deep learning, as well as in RS, have been developed recently. 
Furthermore, \textit{domain adaptation} methods for classification and segmentation have also been developed. 
Models trained on data from one domain may \textit{not} generalize well on other domains. 
Even in one domain, accurate semantic segmentation is challenging. 
There may be a loss in accuracy when deploying a model on \textit{unseen} data due to a shift between the distributions in the source and \textit{target} domains \cite{DomainAdaptationRS}. 
Domain adaptation tries to overcome this \cite{DomainAdversarial, DeepUnsupervised}. 
Domain gaps are common in aerial images \cite{BenchmarkingDomainAdaptation, ModelClassAware}, e.g. region change. 
In \cite{SamePlace}, no off-the-shelf semantic segmentation model transferred well/ accurately on the aerial CVUSA dataset.

\textbf{Architectures.}                    
Several models have been developed recently for semantic segmentation, including \textit{Transformer} (e.g. SegFormer \cite{ModelSegFormer}), encoder-decoder like SegNet or U-Net with skip connections, Fully Convolutional Network (FCN), and dilated convolutions with larger receptive field to capture \textit{long-range} information. 
SegFormer \cite{ModelSegFormer} for semantic segmentation combines the Transformer with an efficient Multi-Layer Perceptron (MLP) decoder and outputs \textit{multi-scale} features.
The decoder combines these multi-scale feature maps, which use local and global attention.
UNetFormer is a UNet-like architecture for segmentation based on a ResNet encoder and a \textit{Transformer} decoder \cite{ModelUNetFormer}. It uses efficient \textit{attention} in the decoder to model global and local information. 
The Segment Anything Model (SAM) \cite{ModelSAM} has also been recently proposed for \textit{instance} segmentation, but not semantic segmentation.

\textbf{Unsupervised adaptation.}                                                            
In the Unsupervised Domain Adaptation (UDA) setting, the model is trained on both labelled and unlabelled data from the source and target domains, respectively. 
Accurate \textit{real-world} semantic segmentation without labels is non-trivial. 
Domain Symmetric Networks (SymNet) \cite{DomainSymmetric, BenchmarkingDomainAdaptation} design the \textit{source and target domains} classifier symmetrically to learn domain-invariant features for effective domain adaptation. 
The model Source Hypothesis Transfer (SHOT) \cite{ModelSHOT} uses hypothesis transfer, training only the backbone and making the classifier of the network \textit{non-trainable}. 
We focus on the scenario in which the number of classes and the classes themselves are the same in the source and target domains, the \textit{closed-set} setting \cite{BenchmarkingDomainAdaptation, UniversalDomainAdaptation}. 
The aim is to achieve good performance during inference on the \textit{target domain} test dataset, as well as on the source domain test dataset \cite{DomainAdversarial, ModelGeoMultiTaskNet}.


\textbf{Semi-supervised learning}                                                                                 
methods for segmentation in deep learning, as well as in RS, have been developed. 
Semi-supervision \cite{SemiSupervisedMiniFrance, ModelFixmatch}, which is halfway between supervised and \textit{unsupervised} learning, deals with settings where labelled sets of data and their targets are provided and unlabelled sets with data \textit{only} are available. 
Unlabelled data help the learning process to \textit{improve} performance. 
To improve generalization \cite{ModelFixmatch, ModelMSMatch}, models should be able to handle labelled and \textit{unlabelled} data, as well as operate within a multi-task optimization framework \cite{SemiSupervisedMiniFrance} performing unsupervised minimization tasks.

\begin{figure}[!tbp]                               
	\centering      \includegraphics[width=0.39\textwidth]{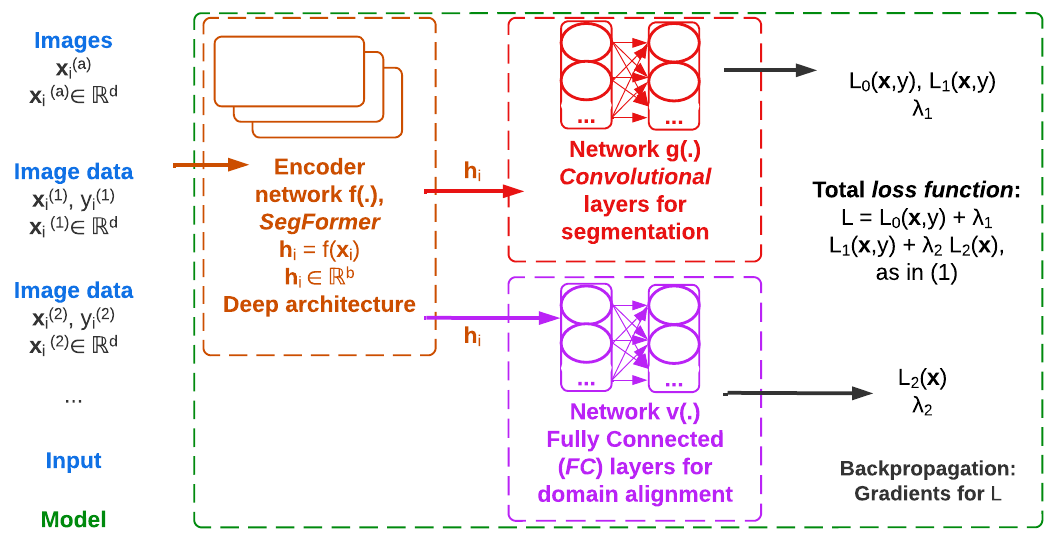}       
	\caption{Flowchart of NEOS for semantic segmentation using domain adaptation on datasets with no \textit{ground truth} labels.}                                                                              
	\label{fig:1}                                                                                
\end{figure}

\section{Proposed Methodology}  
\label{sec:pagestyle}



\textbf{Flowchart.}                               
NEOS is presented in Fig.~\ref{fig:1}. The input is the image from labelled and unlabelled datasets. 
The output is the estimated \textit{semantic} segmentation mask. 
NEOS is based on the SegFormer B5 \cite{ModelSegFormer} architecture and uses a second output for the feature misalignment loss term for domain adaptation. 

\textbf{Loss function.}  NEOS minimizes a loss comprising the terms: (a) cross-entropy for pixel-level classification which is computed using the input labelled images and their corresponding ground truth segmentation masks, (b) ($1$ - \textit{Dice score}) for segmentation, and (c) the \textit{features} misalignment loss. 
The latter is for domain adaptation to enforce the model to be able to work \textit{well} with the different datasets. 
Here, an architecture with \textit{two heads} is used. 
The first two loss terms control the first output head for accurate segmentation and classification on the different \textit{labelled} datasets, while the features misalignment loss, which controls the second output head, enforces the network to reduce the distance between the latent representations of the labelled and unlabelled data in the cross-entropy metric. 
This forces the features to achieve manifold alignment of the embeddings of the different domains. 
We denote the data by $\textbf{x}$ and the ground truth masks by $\textbf{y}$. 
Now, the \textit{cost} function is:\begin{align}                                                           
& \text{argmin}_{f} \ L \text{, } \ \ \ L = L_0(\textbf{x}, \textbf{y}) + \lambda_1 L_1(\textbf{x}, \textbf{y}) + \lambda_2 L_2(\textbf{x})\text{,}   \label{eq:labetioonon0}         
\end{align} 
where we denote our model by $f(\cdot)$, the first, second, and third \textit{loss} terms by $L_0$, $L_1$, and $L_2$, respectively, and the hyperparameters of the second and third losses by $\lambda_1$ and $\lambda_2$.

The first term of our model's objective loss function is: \begin{align}                                          
& L_0 = - \dfrac{1}{N W H} \sum_{j=1}^N \sum_{i=1}^W \sum_{l=1}^H \log \dfrac{\exp(f_{y_{j,i,l}}(\textbf{x}_{j}))}{\sum_{k=1}^K \exp(f_{k,i,l}(\textbf{x}_{j}))}\text{,} \label{eq:labelthelabequatioononn11} 
\end{align}
where $W$ and $H$ are the width and height of the images, $i$ and $l$ the indices for the width and height, $N$ the number of samples, $j$ their index, $K$ the number of classes, and $k$ their index. 
In \eqref{eq:labelthelabequatioononn11}, the $N$ training samples originate from $Q$ labelled datasets. 
This affects both $\textbf{x}_j$ and $y_{j,i,l}$. 
The features before the normalized exponential, i.e. \textit{softmax}, are denoted by $f_{y_{j,i,l}}(\textbf{x}_{j})$. 
The model computes the estimated probability of the labels for the cross-entropy loss to \textit{reward} correct classification, penalizing deviation from the correct class. 
The estimated \textit{semantic} segmentation mask, $\hat{y}_{j,i,l}$ is the pixel-wise class label. 
Here, for the labelled data, to perform accurate classification, NEOS minimizes \eqref{eq:labelthelabequatioononn11}, which is the \textit{pixel-wise} cross-entropy loss.

\begin{figure}[tb]                         
  \centering           
  \centerline{\epsfig{figure=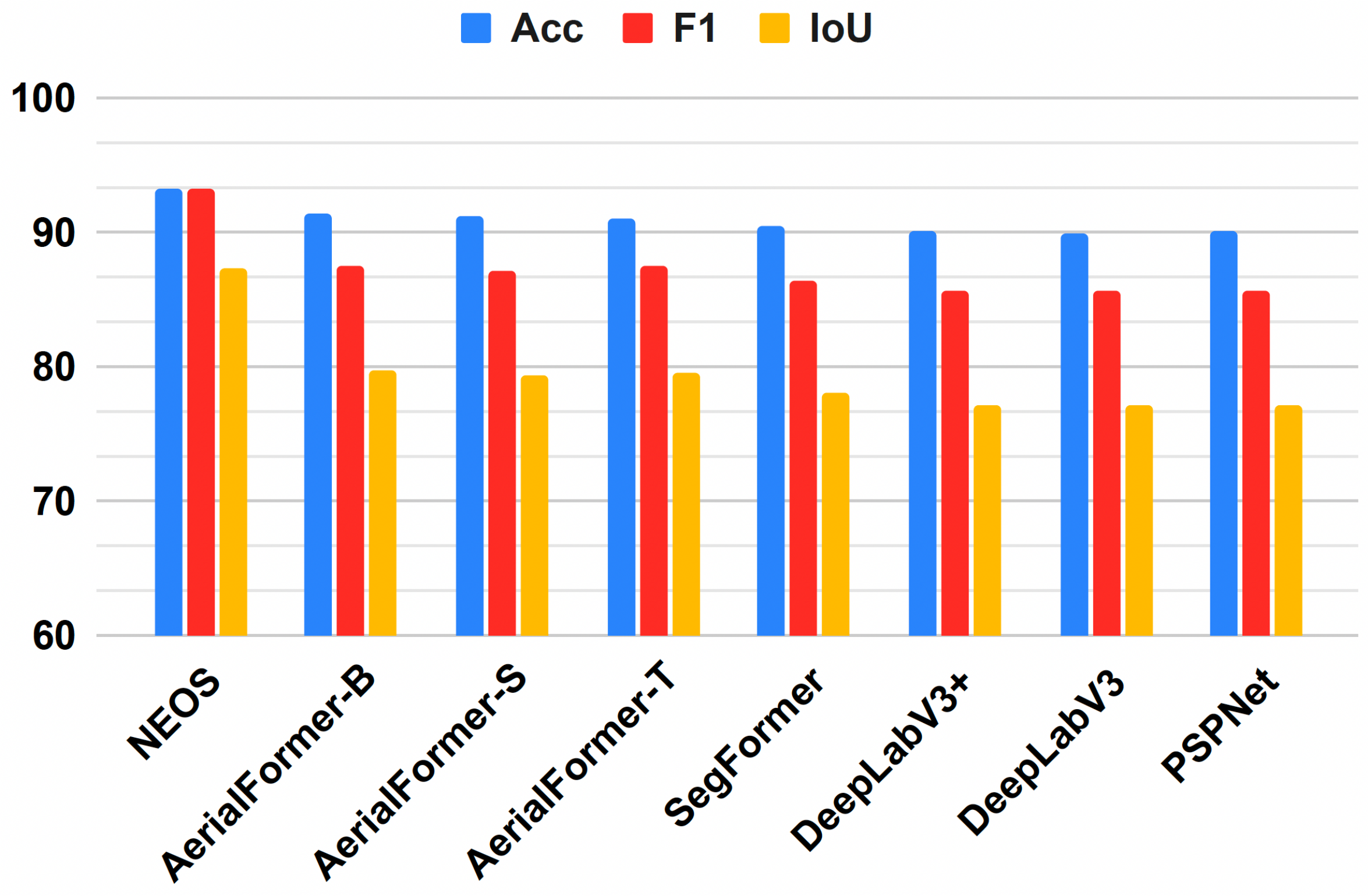,width=7.45cm}}      
\caption{Evaluation of NEOS in accuracy (Acc), F1-score (F1) and IoU on the dataset Potsdam with the class Clutter [12].}                        
\label{fig:resultstousemain1}        
\end{figure}

Next, the second term of the loss function, $L_1$ in \eqref{eq:labetioonon0}, is:

\begin{align}                                                              
& 1 \ - \ \dfrac{2 \ \sum_{j=1}^N \sum_{i=1,l=1}^{W,H} g_{j,i,l} s_{j,i,l}}{\sum_{j=1}^N \sum_{i=1,l=1}^{W,H} g_{j,i,l} + \sum_{j=1}^N \sum_{i=1,l=1}^{W,H} s_{j,i,l}}\text{,} \label{eq:labelththelabequatioonon2}  
\end{align} 
where $g_{j,i,l}$ is the true binary indicator of the class label, and $s_{j,i,l}$ is the estimated \textit{semantic} segmentation probability outputted by the model. 
The $N$ samples originate from $Q$ labelled datasets. 
This affects both $g_{j,i,l}$ and $s_{j,i,l}$. 
To perform accurate segmentation on the labelled data, NEOS minimizes \eqref{eq:labelththelabequatioonon2}, the \textit{Dice} loss.
Next, the third term of the loss function, $L_2$, is:\begin{align}                                             
& L_2 \ \ = \ \ \dfrac{1}{J} \ \sum_{j=1}^J \log \dfrac{\exp(f_{z_{j}}(\textbf{x}_{j}))}{\sum_{m=1}^M \exp(f_{m}(\textbf{x}_{j}))}\text{,} \label{eq:labelthelabequequatioononnn333} 
\end{align}
where the true domain label is denoted by $z_j$, and the number of domain labels by $M$. 
The samples originate from $Q$ labelled and $R$ \textit{unlabelled} datasets. 
For domain adaptation, NEOS minimizes \eqref{eq:labelthelabequequatioononnn333}. 
To improve generalization (better performance), we perform data augmentation to incorporate \textit{invariances} into the model. 
We regularize and enforce the model to generalize and be robust to data transformations, and we also perform downsampling, having inputs at different multi-scale levels.

\section{Evaluation and Results}         
\label{sec:EvaluationResults}

\textbf{Labelled and unlabelled datasets.}                         
We train NEOS on the aerial image datasets: labelled Potsdam and Vaihingen, and unlabelled CVUSA by performing domain adaptation. 
We test NEOS on Potsdam and Vaihingen, as well as on CVUSA. 
We also \textit{compare} our model to other baseline models, including SAM \cite{ModelSAM}. 
For the labelled datasets, the classes are: Buildings (blue colour), Trees (green), Cars (yellow), Low vegetation (cyan), Roads (white), and Clutter (red) \cite{ModelAerialFormer, EmbeddingUNet}. 
Also, for the different datasets and the domain adaptation loss term, as well as the \textit{tags} for the different datasets, Tag A is used for the dataset Potsdam, Tag B for Vaihingen, and C for CVUSA.

\begin{figure}[tb]              
  \centering       
  \centerline{\epsfig{figure=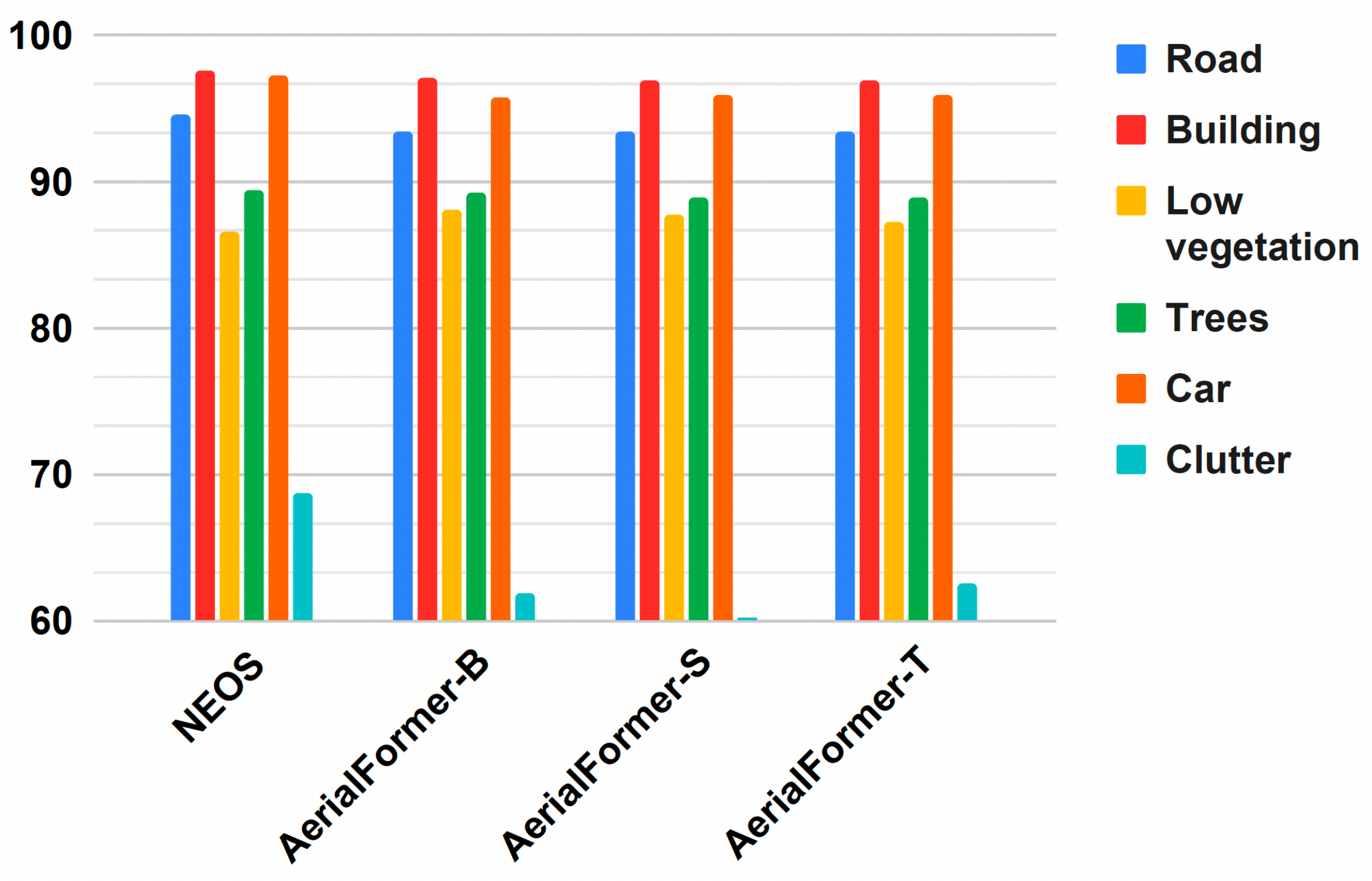,width=0.415\textwidth}}   
\vspace{-9pt}    
\caption{Per-class \textit{F1-score} evaluation (in $\%$) of NEOS on the Potsdam dataset including the class Clutter in the evaluation.}                          
\label{fig:resultstouseuse2}    
\end{figure}

\begin{figure}[tb]                
\hspace{26.25pt}  
\begin{minipage}[b]{.1\linewidth}    
  \centering     
  \centerline{\epsfig{figure=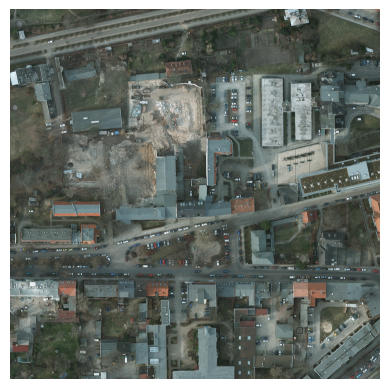,width=3.0cm}}  
  \vspace{0.06cm} 
  \centerline{a) Input}\medskip
\end{minipage}
\hspace{51.25pt}
\begin{minipage}[b]{0.1\linewidth}
  \centering
  \centerline{\epsfig{figure=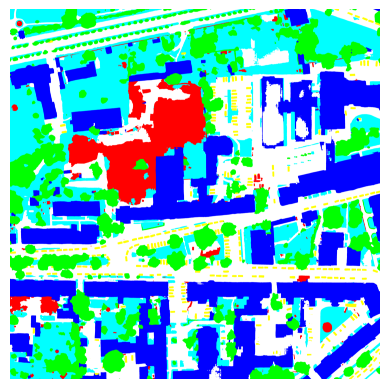,width=3.0cm}} 
  \vspace{0.06cm}
  \centerline{b) NEOS (Ours)}\medskip 
\end{minipage}
\hspace{51.25pt}  
\begin{minipage}[b]{.1\linewidth}  
  \centering  
  \centerline{\epsfig{figure=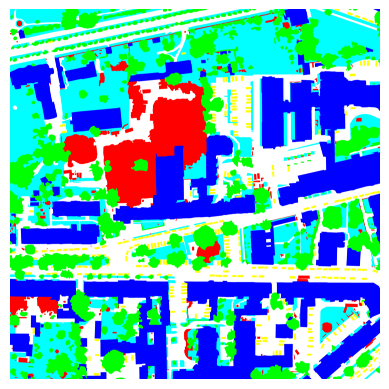,width=3.0cm}}
  \vspace{0.06cm}
  \centerline{c) Ground truth}\medskip
\end{minipage}               
\vspace{-9pt}      
\caption{Semantic segmentation masks by NEOS on Potsdam.}                      
\label{fig:resultsPotsdamNEOS}    
\end{figure}

\textbf{Evaluation of NEOS on Potsdam.}                                                                   
We evaluate NEOS on the dataset Potsdam in Fig.~\ref{fig:resultstousemain1}. 
In this experiment, the class Clutter is considered in the evaluation. 
Here, the evaluation is based on the accuracy, F1-score, and Intersection over Union (IoU) metrics. 
It can be observed in Fig.~\ref{fig:resultstousemain1} that the proposed model outperforms all the other baseline models \cite{ModelAerialFormer}.
In Fig.~\ref{fig:resultstouseuse2}, we evaluate the \textit{per-class} F1-score performance of NEOS on Potsdam. 
In Fig.~\ref{fig:resultsPotsdamNEOS}, we present NEOS \textit{qualitative} results.

\begin{figure}[tb]                           
  \centering             
  \centerline{\epsfig{figure=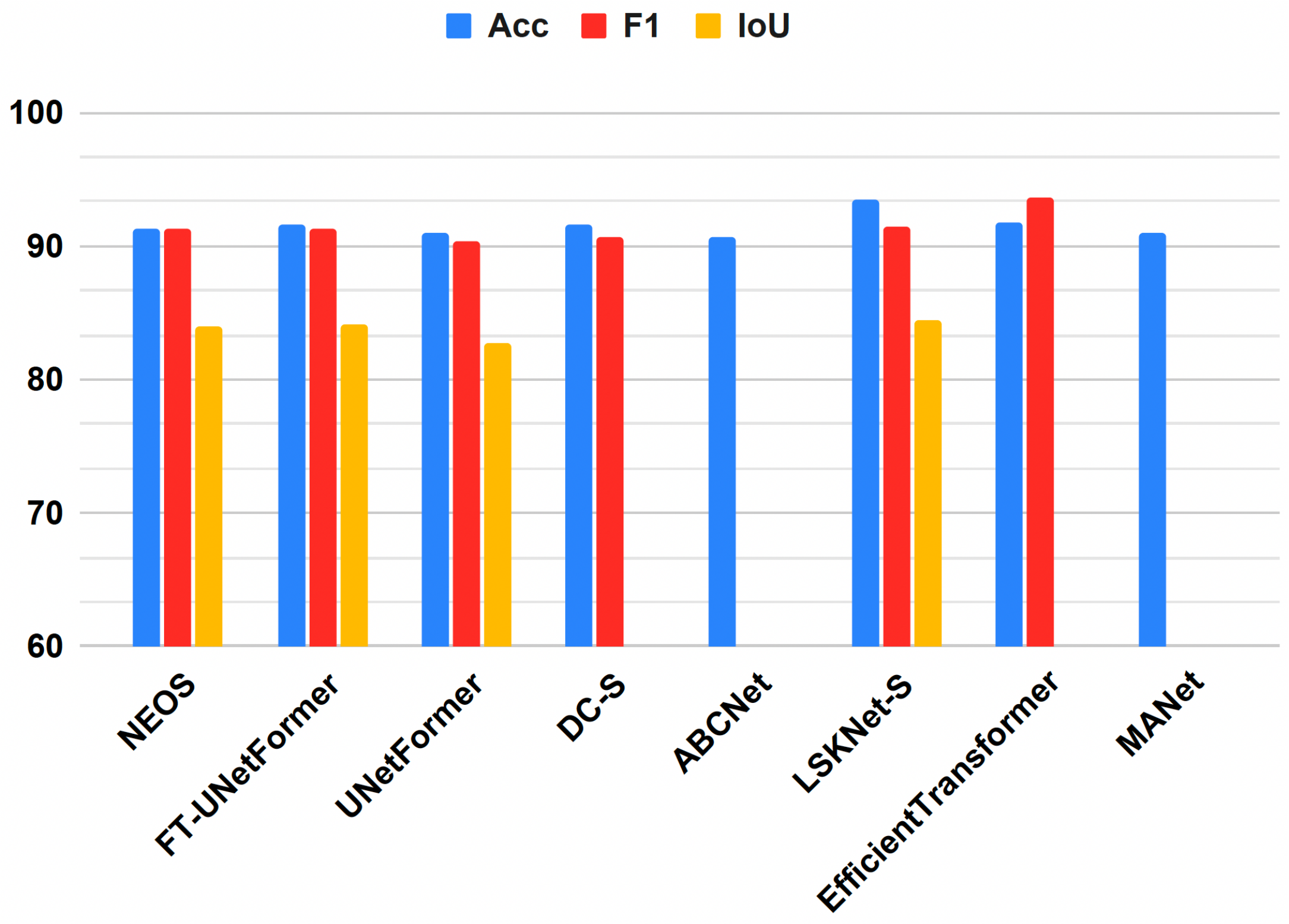,width=0.4125\textwidth}}     
\vspace{-9pt}     
\caption{Evaluation of NEOS on Vaihingen in Acc, F1 and IoU.}                              
\label{fig:resultstousee25}      
\end{figure}

\begin{figure}[tb]                   
  \centering       
  \centerline{\epsfig{figure=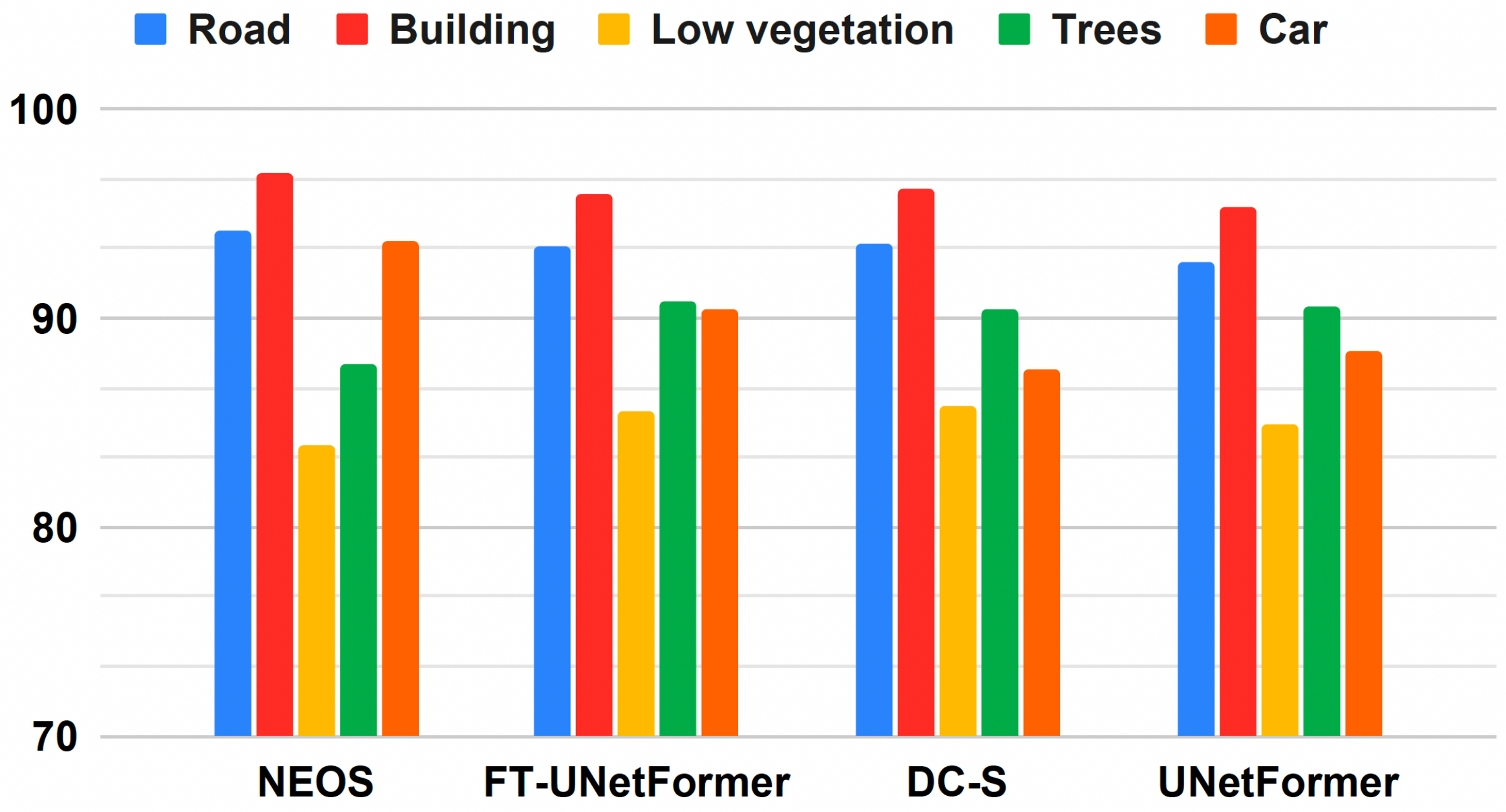,width=0.38\textwidth}}      
\vspace{-9pt}     
\caption{Per-class F1-score evaluation of NEOS on Vaihingen.}                          
\label{fig:resultstotouse2}      
\end{figure}

\textbf{Evaluation of NEOS on Vaihingen.}                                    
We evaluate NEOS in Fig.~\ref{fig:resultstousee25} in accuracy, F1-score and IoU. 
Here, the results of NEOS on Vaihingen are comparable to other models. 
In Fig.~\ref{fig:resultstotouse2}, we also evaluate the \textit{F1-score} of NEOS for each class \cite{ModelUNetFormer}. 
For Roads, Buildings and Cars, NEOS outperforms other models.

\begin{figure}[tb]                  
\hspace{26pt}  
\begin{minipage}[b]{.1\linewidth}      
  \centering     
  \centerline{\epsfig{figure=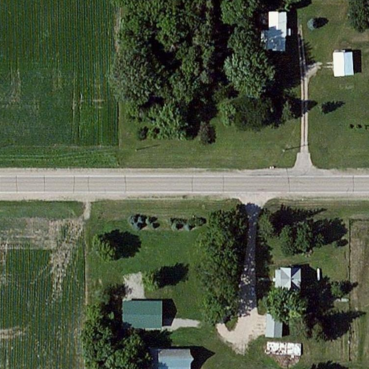,width=2.93cm}}  
  \vspace{0.06cm} 
  \centerline{a) Input}\medskip
\end{minipage}
\hspace{49pt}
\begin{minipage}[b]{0.1\linewidth}
  \centering
  \centerline{\epsfig{figure=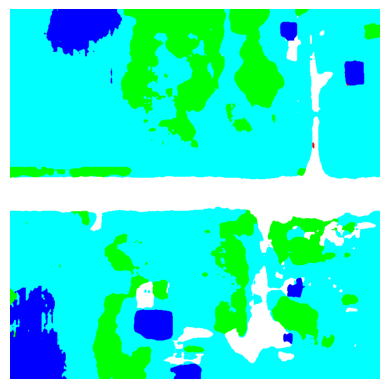,width=3.0cm}} 
  \vspace{0.06cm}
  \centerline{b) NEOS (Ours)}\medskip  
\end{minipage}
\hspace{55pt}    
\begin{minipage}[b]{.1\linewidth}  
  \centering  
  \centerline{\epsfig{figure=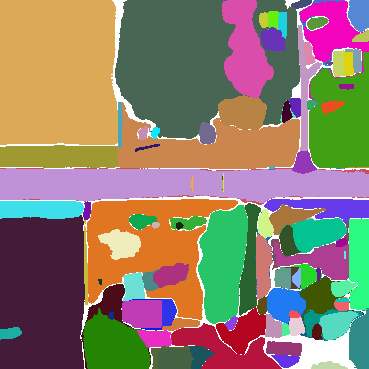,width=2.925cm}}   
  \vspace{0.06cm}
  \centerline{c) samgeoHQ}\medskip 
\end{minipage}              

\hspace{26pt}  
\begin{minipage}[b]{.1\linewidth}        
  \centering      
  \centerline{\epsfig{figure=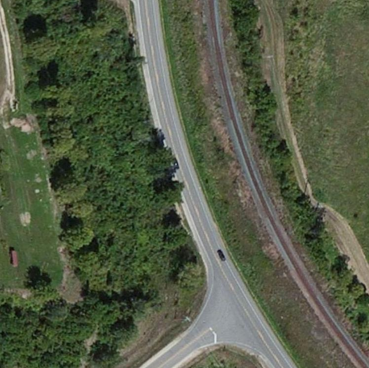,width=2.93cm}}  
  \vspace{0.06cm}  
  \centerline{d) Input}\medskip
\end{minipage}
\hspace{49pt}
\begin{minipage}[b]{0.1\linewidth} 
  \centering 
  \centerline{\epsfig{figure=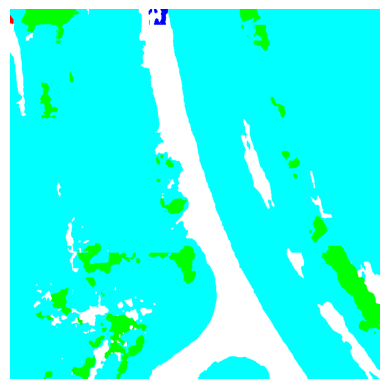,width=3.0cm}} 
  \vspace{0.06cm}
  \centerline{e) NEOS (Ours)}\medskip  
\end{minipage}
\hspace{55pt}    
\begin{minipage}[b]{.1\linewidth}   
  \centering    
  \centerline{\epsfig{figure=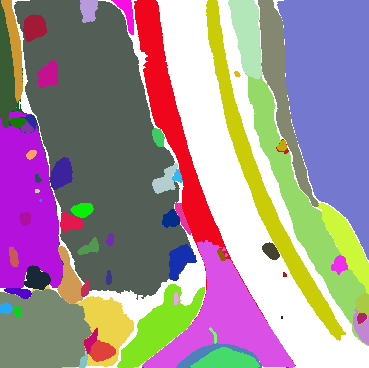,width=2.925cm}}    
  \vspace{0.06cm} 
  \centerline{f) samgeoHQ}\medskip       
\end{minipage}  
\vspace{-9pt}      
\caption{Qualitative evaluation of NEOS on the unlabelled CVUSA aerial dataset, and comparison to samgeoHQ \cite{ModelSAMGEO}.}                           
\label{fig:res4imimagesNEOSS}    
\end{figure}

\textbf{Evaluation on the unlabelled dataset CVUSA.}                                                                                     
We evaluate NEOS on the \textit{unlabelled} dataset aerial CVUSA. 
The testing is done on a dataset that does \textit{not} have ground truth masks. 
In Fig.~\ref{fig:res4imimagesNEOSS}, we present the \textit{qualitative} results of NEOS. 
We observe in (b) and (e) that NEOS performs semantic segmentation and is able to \textit{recognize} effectively classes such as Roads and Low vegetation. 
This holds for NEOS for the vertical roads that have \textit{shadows} (occlusion) in (a)-(b). 
In the next paragraphs, we present a numerical evaluation of NEOS on the unlabelled CVUSA dataset and a further comparison to other models.

In Fig.~\ref{fig:res4imimagesNEOSS}, we examine the qualitative results of NEOS, and we also do this at \textit{large scale}, automating the process. 
We assess the performance of NEOS on many images, and to capture the big picture, we evaluate NEOS \textit{numerically} by computing the Segments of Predictions and Inputs Error (SPIE). 
We first perform detection of segments on the estimated mask and input images, and for this, we use a variant of SAM \cite{ModelSAMEO}. 
Then, the error is calculated and normalized. 
For perfect segmentation without considering semantic information (no labels), SPIE is \textit{zero}.  
For completely inaccurate segmentation, SPIE is equal to one. 
SPIE is the mean residual image where the residual is between the \textit{estimated} mask and the input after being modified by a detection of segments algorithm, and its definition is:\begin{align}     
& \text{SPIE} \ \ = \ \ \dfrac{1}{R} \ \sum_{j=1}^R g(f(\textbf{x}_j)) - g(\textbf{x}_j) \label{eq:labetioononlast}           
\end{align}where $f(\cdot)$ is NEOS from \eqref{eq:labetioonon0} (or another model), $R$ the number of evaluation samples, and $g(\cdot)$ a detection of segments algorithm. 
We use SPIE as an indicator of good performance and as an empirical metric that works in practice.   
Using SPIE in \eqref{eq:labetioononlast}, the numerical results \textit{match} the qualitative results we obtain. 
In addition, we have also included the code in \cite{ModelNEOS}.

\begin{table}[!tb]           
    \caption{Evaluation of NEOS on the CVUSA dataset, on both the aerial (Aer) and street (Str), and the improvement (I) over the base model.}\label{tab:table1results}   
      \vspace{5pt} \centering       
        \begin{tabular}
{p{3.22cm} p{0.7cm} p{0.7cm} p{0.7cm} p{0.65cm}}  
\toprule    
    \normalsize {\small \textbf{SPIE for aerial \& street}} & \normalsize {\small \textbf{Aer}} & \normalsize {\small \textbf{I Aer}} & \normalsize {\small \textbf{Str}} & \normalsize {\small \textbf{I Str}}
\\
\midrule 
\midrule
\normalsize {\small NEOS (Ours)}
& \normalsize {\normalsize  \small $0.047$} & 
    \normalsize {\normalsize \small $32\%$} &   
    \normalsize {\normalsize \small $0.041$} & 
    \normalsize {\normalsize \small $21\%$}   
\\   
\midrule 
\midrule
\normalsize {\small Base model, SegFormer} & \normalsize {\normalsize  \small $0.069$} &
\normalsize {\normalsize  \small $\text{N/A}$} &
\normalsize {\normalsize  \small $0.052$} &  
\normalsize {\normalsize  \small $\text{N/A}$} 
\\ 
\midrule 
\normalsize {\small CNN-based using Eq.~\eqref{eq:labetioonon0}} &  \normalsize {\normalsize  \small $0.064$} &
\normalsize {\normalsize  \small $7.2\%$} & 
\normalsize {\normalsize  \small $0.049$} &
\normalsize {\normalsize  \small $5.8\%$} 
\\ 
\midrule           
\midrule  
\end{tabular}          
\end{table}

We evaluate NEOS numerically on CVUSA in Table~\ref{tab:table1results} using SPIE, and we compare with other models for \textit{semantic} segmentation.
The improvement of NEOS for aerial images over the base model, SegFormer \cite{ModelSegFormer}, is $32\%$. 
NEOS also outperforms \cite{ModelSegFormer} for street images when we use the labelled dataset CityScapes \cite{DatasetCityScapes} for the domain adaptation loss in Sec.~\ref{sec:pagestyle}.

\textbf{Further comparison of NEOS to other models.}                                                       
In Fig.~\ref{fig:res4imimagesNEOSS}, we examine the results of NEOS and its comparison to SAM \cite{ModelSAM}. 
NEOS performs \textit{joint} classification and segmentation, while SAM does \textit{not} consider semantics \cite{SAMRobustness}. 
There are several \textit{variants} of SAM: 
Geospatial SAM \cite{ModelSAMGEO} \textit{fine-tunes} \cite{ModelSAM} on aerial data. 
SAM-HQ \cite{ModelSAMHQ} achieves improved accuracy (FastSAM \cite{ModelFastSAM}, speed). 
Semantic SAM (SSAM) \cite{ModelSSAM, ModelSSAM2} modifies \cite{ModelSAM} to perform semantic (rather than instance) segmentation, but \textit{not} for aerial data. 
We examined the performance of SAM and its variants on the unlabelled CVUSA dataset. 
In Fig.~\ref{fig:res4imimagesNEOSS}, samgeoHQ \cite{ModelSAMGEO}, which does \textit{not} perform semantic segmentation, is sensitive to even small changes in the scene. 
For scenes with details, we need to adjust the several tunable parameters of SAM to control how \textit{dense} the estimated masks are. 
For NEOS, we do not need to adjust its parameters, and this can potentially lead to improved user convenience and ease of use.

\section{Conclusion}        
\label{sec:majhead}  

We have developed a semantic segmentation method that is effective for unlabelled datasets. 
The results show that NEOS outperforms other models. 
We have used the unlabelled aerial CVUSA dataset, where accurate semantic segmentation has not yet been performed, to the best of the authors' knowledge, and we plan to also use the results for cross-view geo-location matching to accurately match aerial and street images \cite{FrancescoProPaper, MastersThesisFrancesco}.

\bibliographystyle{IEEEbib}  
\bibliography{strings}

\end{document}